\documentclass[sigconf]{acmart}

\usepackage{times}

\usepackage{hyperref}
\usepackage{url}
\usepackage{xcolor}

\long\def\ignore#1{}

\title{To Tune or Not To Tune?\\ Zero-shot Models for Legal Case Entailment}

\AtBeginDocument{%
  \providecommand\BibTeX{{%
    \normalfont B\kern-0.5em{\scshape i\kern-0.25em b}\kern-0.8em\TeX}}}

\copyrightyear{2021}
\acmYear{2021}
\setcopyright{acmcopyright}\acmConference[ICAIL'21]{Eighteenth International
Conference for Artificial Intelligence and Law}{June 21--25, 2021}{São Paulo, Brazil}
\acmBooktitle{Eighteenth International Conference for Artificial Intelligence and Law
(ICAIL'21), June 21--25, 2021, São Paulo, Brazil}
\acmPrice{15.00}
\acmDOI{10.1145/3462757.3466103}
\acmISBN{978-1-4503-8526-8/21/06}


\begin{document}

\author{Guilherme Moraes Rosa}
\affiliation{%
  \institution{NeuralMind}
  \institution{University of Campinas (Unicamp)}
  \country{Brazil}
}

\author{Ruan Chaves Rodrigues}
\affiliation{
  \institution{NeuralMind}
  \institution{Federal University of Goiás (UFG)}
  \country{Brazil}
}

\author{Roberto de Alencar Lotufo}
\affiliation{%
  \institution{NeuralMind}
  \institution{University of Campinas (Unicamp)}
  \country{Brazil}
}

\author{Rodrigo Nogueira}
\affiliation{%
  \institution{NeuralMind}
  \institution{University of Campinas (Unicamp)}
  \institution{University of Waterloo, Canada}
  \country{}
}

\renewcommand{\shortauthors}{Rosa et al.}
\renewcommand{\shorttitle}{Zero-shot Models for Legal Case Entailment}

\begin{abstract}
There has been mounting evidence that pretrained language models fine-tuned on large and diverse supervised datasets can transfer well to a variety of out-of-domain tasks.
In this work, we investigate this transfer ability to the legal domain.
For that, we participated in the legal case entailment task of COLIEE 2021, in which we use such models with no adaptations to the target domain.
Our submissions achieved the highest scores, surpassing the second-best team by more than six percentage points.
Our experiments confirm a counter-intuitive result in the new paradigm of pretrained language models:\ given limited labeled data, models with little or no adaptation to the target task can be more robust to changes in the data distribution than models fine-tuned on it.
Code is available at~\url{https://github.com/neuralmind-ai/coliee}.
\end{abstract}

\maketitle

\section{Introduction}

An ongoing trend in natural language processing and information retrieval is to use the same model with small adaptations to solve a variety of tasks.
Pretrained transformers, epitomized by BERT~\cite{devlin2019bert}, are the state-of-the-art in question answering~\cite{khashabi2020unifiedqa}, natural language inference~\cite{he2020deberta,raffel2020t5}, summarization~\cite{lewis2020bart,bao2020unilmv}, and ranking tasks~\cite{ma2021prop,gao2021rethink}.
Although these tasks are diverse, current top-performing models in each of them have a similar architecture to the original Transformer~\cite{vaswani2017attention} and are pretrained on variations of the masked language modeling objective used by \citet{devlin2019bert}.

Zero-shot and few-shot models are becoming more competitive with models fine-tuned on large datasets.
For instance, few-shot results of GPT-3~\cite{NEURIPS2020_1457c0d6} sparked an interest in prompt engineering methods, which are now an active area of research~\cite{schick2020exploiting,lu2020learning,tam2021improving}.
The goal of these methods is to find input templates such that the model is more likely to give the correct answer.

In information retrieval, pretrained models fine-tuned only on a large dataset have also shown strong zero-shot capabilities~\cite{thakur2021beir}.
For example, the same multi-stage pipeline based on T5~\cite{raffel2020t5} was the best or second best-performing system in 4 tracks of the TREC 2021~\cite{pradeep5h2oloo}, including specialized tasks such as Precision Medicine~\cite{Roberts2019OverviewOT}, and TREC-COVID~\cite{zhang2020rapidly}.
A remarkable feature of this pipeline is that, for most tasks, the models are fine-tuned only on a general-domain ranking dataset, i.e., they do not use in-domain data.


However, to date, there has not been strong evidence that zero-shot models transfer well to the legal domain.
Most state-of-the-art models need adaptations to the target task.
For example, the top-performing system on the legal case entailment task of COLIEE 2020~\cite{rabelo2020coliee} uses an interpolation of BM25~\cite{robertson1995okapi} scores and scores from a BERT model fine-tuned on the target task~\cite{nguyen2020jnlp}.

In this work, we show that, for the legal case entailment task of COLIEE, pretrained language models without any fine-tuning on the target task perform at least equivalently or even better than models fine-tuned on the task itself.
Our approach is characterized as zero-shot since the model was only fine-tuned on annotated data from another domain.
Our result confirms in the legal domain a counter-intuitive recent finding in other domains:\ given limited labeled data, zero-shot models tend to perform better on held-out datasets than models fine-tuned on the target task~\cite{radford2021learning,pradeep5h2oloo}.


\section{Related Work}


It is a common assumption among NLP researchers that models developed using nonlegal texts would lead to unsatisfactory performance when directly applied to legal tasks~\cite{zhong2020, Elnnagar1}. To overcome this issue, general-purpose techniques are adapted to the legal domain. For example, \citet{Chalkidis2019} pre-trained legal word embeddings using word2vec~\cite{word2vec1, word2vec2} over a large corpus comprised of legislation from multiple countries. \citet{zhong2020b} created a question answering dataset in the legal domain, collected from the National Judicial Examination of China and evaluated different models on it, including Transformers. \citet{Elnnagar2} applied multi-task learning to minimize the problems related to data scarcity in the legal domain. The models were trained in translation, summarization, and multi-label classification tasks, and achieved better results than single-task models. 

Pretrained transformer models have only begun to be adopted in legal NLP applications more broadly~\citep{yeung2019effects,elwany2019bert,shaghaghian2020customizing,leivaditi2020benchmark,bambroo2021legaldb}.
In some tasks, they marginally outperform classical methods, especially when training data is scarce.
For example, \citet{zhong2020does} showed that a BERT-based model performs better than a tf-idf similarity model on a judgment prediction task~\citep{xiao2018cail2018}, but is slightly less effective than an attention-based convolutional neural network~\citep{yin2016abcnn}.

In some cases, they outperform classical methods, but at the expense of using hand-crafted features or by being fine-tuned on the target task. For example, the best submission to task 2 of COLIEE 2019 was a BERT model fed with hand-crafted inputs and fine-tuned on in-domain data~\citep{rabelo2019combining}.

\citet{peters2019tune} demonstrate that fine-tuning on the target task may not perform better than simple feature extraction from a pretrained model if the pretraining task and the target task belong to highly different domains. These findings lead us to consider zero-shot approaches while investigating how general domain Transformer models can be applied to legal tasks.

Although zero-shot approaches are relatively novel in the legal domain, our work is not the first to apply zero-shot Transformer models to domain-specific entailment tasks where
limited labeled data is available. Yin et al. \cite{yin2019benchmarking} have transformed multi-label classification tasks into textual entailment tasks, and then evaluated the performance of a BERT model fine-tuned on mainstream entailment datasets. Yin et al. \cite{yin2020universal} also performed similar experiments while transforming question answering and coreference resolution tasks into entailment tasks.
We are not the first to use zero-shot techniques on the legal case entailment task. For instance, \citet{rabelo2020} used a BERT fine-tuned for paraphrase detection combined with two transformer-based models fine-tuned on a generic text entailment dataset and features generated by a BERT model fine-tuned on the COLIEE training dataset.
However, we are the first to show that zero-shot models can outperform fine-tuned ones on this task.

\subsection{The Legal Case Entailment Task}

The Competition on Legal Information Extraction/Entailment (COLIEE)~\citep{rabelo2020coliee,rabelo2019coliee,kano2018coliee,Kano2017OverviewOC} is an annual competition whose aim is to evaluate automatic systems on case and statute law tasks.

Among the five tasks of the 2021 competition, we submitted systems to task 2, called legal case entailment, which consists of identifying paragraphs from existing cases that entail a given fragment of a base case.

\begin{figure*}[h]
  \centering
  \includegraphics[width=17cm]{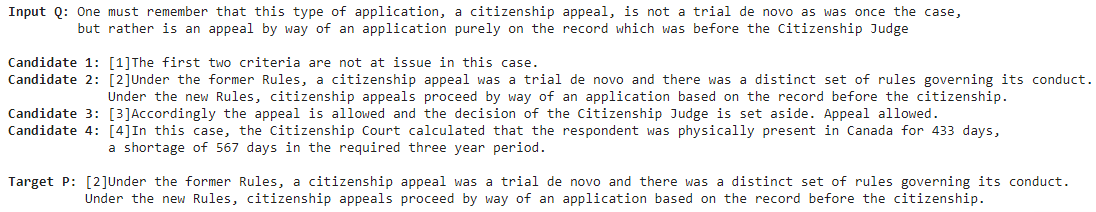}
  \caption{COLIEE's Task 2 example.}
  \label{fig:coliee}
\end{figure*}

Training data consists of a set of decision fragments, its respective candidate paragraphs that could be relevant or not to the fragment and a set of labels containing the number of the paragraphs by which the decision fragment is entailed. Test data includes only decision fragments and candidate paragraphs, but no labels. As shown in Figure \ref{fig:coliee}, the input to the model is a decision fragment Q of an unseen case and the output should be a set of paragraphs $P = [P_1, P_2, ..., P_n]$ that are relevant to the given decision $Q$. In table~\ref{table:task2}, we show the statistics of the 2020 and 2021 datasets.

We separate 80\% of the 2020 training set for training and the remaining for validation, which yields 260 and 65 positive examples for training and validation sets, respectively.
Negative examples are all candidates not labeled as positive.

\begin{table}[h!]

\centering\centering\resizebox{0.48\textwidth}{!}{
 \begin{tabular}{l | c | c | c | c } 
 \toprule
 & \multicolumn{2}{c|}{\textbf{2020}} & \multicolumn{2}{c}{\textbf{2021}} \\ 
 & \textbf{Train} &  \textbf{Test} & \textbf{Train} &  \textbf{Test}  \\
 \midrule
 Examples (base cases) & 325 & 100 & 425 & 100  \\ 
 
 Avg. \# of candidates / example & 35.52 & 36.72 & 35.80 & 35.24 \\
 
 Avg. positive candidates / example & 1.15 & 1.25 & 1.17 & 1.17 \\
 
 Avg. of tokens in base cases & 37.72 & 37.03 & 37.51 & 32.97 \\
 Avg. of tokens in candidates & 100.16 & 112.65 & 103.14 & 100.83 \\
 \bottomrule
\end{tabular}
}
\vspace{0.1cm}
\caption{Statistics of COLIEE's Task 2.}
\label{table:task2}
\end{table}


The micro F1-score is the official metric in this task:
\begin{equation}
    \text{F1} = (2 \times P \times R) / (P + R),
\end{equation}
\noindent where $P$ is the number of correctly retrieved paragraphs for all queries divided by the number of retrieved paragraphs for all queries, and $R$ is the number of correctly retrieved paragraphs for all queries divided by the number of relevant paragraphs for all queries.

\section{Method}

We experiment with the following models: BM25, monoT5-zero-shot, monoT5, and DeBERTa. We also evaluate an ensemble of our monoT5 and DeBERTa models.

\subsection{BM25}
BM25 is a bag-of-words retrieval function that scores a document based on the query terms appearing in it.
We use the BM25 implemented in Pyserini~\cite{lin2021pyserini}, a Python toolkit that supports replicable information retrieval research.
We use its default parameters.

We first index all paragraphs in datasets of tasks 1 and 2.
Having more paragraphs from task 1 improves the term statistics (e.g., document frequencies) used by BM25. Task 1 dataset is composed of long documents, while task 2 is composed of paragraphs. This difference in length may degrade BM25 scores for task 2 paragraphs because the average document length will be higher due to task 1 documents. We address this problem by segmenting each document into several paragraphs using a context window of 10 sentences with overlapping strides of 5 sentences.

The entailed fragment might be comprised of multiple sentences. Here we treat each of its sentences as a query and compute a BM25 score for each sentence and candidate paragraph pair independently. The final score for each paragraph is the maximum among its sentence and paragraph pair scores.
We then use the method described in Section~\ref{section:answer_selection} to select the paragraphs that will comprise our final answer.

\subsection{monoT5-zero-shot}

At a high level, monoT5-zero-shot is a sequence-to-sequence adaptation of the T5 model~\cite{raffel2020t5} proposed by~\citet{nogueira2020document} and further detailed in~\citet{lin2020pretrained}.
This ranking model is close to or at the state-of-the-art in retrieval tasks such as Robust04 \cite{trec2004}, TREC-COVID, and TREC 2020 Precision Medicine and Deep Learning tracks.
Details of the model are described in~\citet{nogueira2020document}; here, we only provide a short overview.

In the T5 model, all target tasks are cast as sequence-to-sequence tasks.
For our task, we use the following input sequence template:
\begin{equation}
\text{Query: } q \ \ \text{ Document: } d \ \ \text{ Relevant:}
\end{equation}

\noindent where $q$ and $d$ are the query and candidate texts, respectively.
In this work, $q$ is a fragment, and $d$ is one of the candidate paragraphs.

The model estimates a score $s$ quantifying how relevant a candidate text $d$ is to a query $q$.
That is:
\begin{equation}
s = P(\textrm{Relevant}=1 | d, q).
\end{equation}

The model is fine-tuned to produce the tokens ``true'' or ``false'' depending on whether the candidate is relevant or not to the query.
That is, ``true'' and ``false'' are the ``target tokens'' (i.e., ground truth predictions in the sequence-to-sequence transformation).
The suffix ``Relevant:'' in the input string serves as hint to the model for the tokens it should produce.

We use a T5-large model fine-tuned on MS MARCO~\cite{MS_MARCO_v3}, a dataset of approximately 530k query and relevant passage pairs.
We use a checkpoint available at Huggingface's model hub that was trained with a learning rate of $10^{-3}$ using batches of 128 examples for 10k steps, or approximately one epoch of the MS MARCO dataset.\footnote{https://huggingface.co/castorini/monot5-large-msmarco-10k}
In each batch, a roughly equal number of positive and negative examples is sampled.
We refer to this model as monoT5-zero-shot.

Although fine-tuning for more epochs leads to better performance on the MS MARCO development set, \citet{nogueira2020document} showed that further training degrades a model's zero-shot performance on other datasets.
We observed similar behavior in our task and opted to use the model trained for one epoch on MS MARCO.

At inference time, to compute probabilities for each query-candidate pair, a softmax is applied only on the logits of the tokens ``true'' and ``false''.
The final score of each candidate is the probability assigned to the token ``true''.

\subsection{monoT5}
We further fine-tune monoT5-zero-shot on the 2020 task 2 training set following a similar training procedure described in the previous section.

Fragments are mostly comprised of only one sentence, while candidate paragraphs are longer, sometimes exceeding 512 tokens in length.
Thus, to avoid excessive memory usage due to the quadratic memory cost of Transformers with respect to the sequence length, we truncate inputs to 512 tokens during both training and inference.

The model is fine-tuned with a learning rate of $10^{-3}$ for 80 steps using batches of size 128, which corresponds to 20 epochs.
Each batch has the same amount of positive and negative examples.
We refer to this model as monoT5.

\begin{table*}[ht]

\centering
 \begin{tabular}{l | l | c | c | c | c | c | c | c } 
 \toprule
 & & \multicolumn{3}{c|}{\textbf{2020}} & \multicolumn{3}{c}{\textbf{2021}}  \\ 
\textbf{Description} & \textbf{Submission name} & \textbf{F1} &  \textbf{Prec} &  \textbf{Recall} & \textbf{F1} &  \textbf{Prec} &  \textbf{Recall} & $\alpha, \beta, \gamma$  \\
 \midrule
 (1a) Median of submissions & - & 0.5718 & - & - & 0.5860 & - & - & -  \\
 (1b) Best of 2020~\cite{nguyen2020jnlp}  & JNLP.task2.BMWT & 0.6753 & 0.7358 & 0.6240 & - & - & - & - \\
 (1c) 2nd best of 2021  & UA\_reg\_pp & - & - & - & 0.6274 & - & - & - \\
 \midrule
 (2) BM25 & - & 0.6046 & 0.7222  & 0.52 & 0.6009 & 0.6666 & 0.5470 & 0.07, 2, 0.99 \\
 (3) DeBERTa & DeBERTa & \textbf{0.7094} & 0.7614 & 0.6640 & 0.6339 & 0.6635 & 0.6068 & 0, 2, 0.999 \\ 
 (4) monoT5 & monoT5 & 0.6887 & 0.7155 & 0.660 & 0.6610 & 0.6554 & 0.6666 & 0, 3, 0.995 \\
 (5) monoT5-zero-shot & - & 0.6577 & 0.7400 & 0.5920  & \textbf{0.6872} & 0.7090 & 0.6666 & 0, 3, 0.995 \\
 \midrule
 (6) Ensemble of (3) and (4) & DebertaT5 & 0.7217 & 0.7904 & 0.6640 & 0.6912 & 0.7500 & 0.6410 & 0.6, 2, 0.999  \\
 (7) Ensemble of (3) and (5) & - & 0.7038 & 0.7592 & 0.6560 & 0.6814 & 0.7064 & 0.6581 & 0.6, 2, 0.999 \\
 \bottomrule
\end{tabular}
\vspace{0.1cm}
\caption{\label{tab:task_2} Test set results on Task 2 of COLIEE 2020 and 2021. Our best single model F1 for each year is in bold.}
\label{table:main_results}
\end{table*}

\subsection{DeBERTa} \textit{Decoding-enhanced BERT with disentangled attention} (DeBERTa) improves on the original BERT and RoBERTa architectures by introducing two techniques: the disentangled attention mechanism and an enhanced mask decoder \cite{he2020deberta}. Both improvements seek to introduce positional information to the pretraining procedure, both in terms of the absolute position of a token and the relative position between them.

The COLIEE 2021 Task 2 dataset has very few positive examples of entailment. Therefore, for fine-tuning DeBERTa on this dataset, we found appropriate to artificially expand the positive examples. As fragments take up only a small portion of a base case paragraph, we expand positive examples by generating artificial fragments from the same base case paragraph in which the original fragment has occurred. This is done by moving a sliding window, with a stride that is half the size of the original fragment, over the base case paragraph. Each step of this sliding window is taken to be an artificial fragment, and such artificial fragments are assigned the same labels as the original fragment.

Although the resulting dataset after these operations is several times larger than the original Task 2 dataset, we achieved better results by fine-tuning DeBERTa on a small sample taken from this artificial dataset. After experimenting with distinct sample sizes, we settled for a sample of twenty thousand fragment and candidate paragraph pairs, equally balanced between positive and negative entailment pairs.

In order to find the best hyperparameters for fine-tuning a DeBERTa Large model, we perform a grid search over the hyperparameters suggested by He et al. \cite{he2020deberta} while early stopping always at the second epoch. The best combination of hyperparameters is used to fine-tune the model for ten epochs. The checkpoint with the best performance on the 2020 test set is selected to generate our predictions for the 2021 test set.

\subsection{Answer Selection} 
\label{section:answer_selection}

The models described above estimate a score for each (fragment, candidate paragraph) pair.
To select the final set of paragraphs for a given fragment, we apply three rules:
\begin{itemize}
    \item Select paragraphs whose scores are above a threshold $\alpha$;
    \item Select the top $\beta$ paragraphs with respect to their scores;
    \item Select paragraphs whose scores are at least $\gamma$ of the top score.
\end{itemize}

\noindent We use exhaustive grid search to find the best values for $\alpha$, $\beta$, $\gamma$ on the development set of the 2020 task 2 dataset. 
We swept $\alpha = [0, 0.1, ..., 0.9]$, $\beta = [1, 2 ..., 10]$, and $\gamma = [0, 0.1, ..., 0.9, 0.95, 0.99, 0.995, \\ 
...,  0.9999]$.
The best values for each model can be found in Table~\ref{table:ablation}.
Note that our hyperparameter search includes the possibility of not using the first or third strategies if $\alpha=0$ or $\gamma=0$ are chosen, respectively.

\subsection{DeBERTa + monoT5 Ensemble (DebertaT5)}
 
Ensemble methods seek to combine the strengths and compensate for the weaknesses of the models in order that the final model has better generalization performance.

We use the following method to combine the predictions of monoT5 and DeBERTa (both fine-tuned on COLIEE 2020):\ We concatenate the final set of paragraphs selected by each model. We remove duplicates, preserving the highest score. Then, we apply again the grid search method explained in the previous section to select the final set of paragraphs. It is important to note that our method does not combine scores between models. It ensures that only individual answers with a certain degree of confidence are maintained in the final answer, which generally leads to an increase in Precision. The final answer for each test example can be composed of individual answers from one model or both models.

\section{Results}

We present our main result in Table~\ref{table:main_results}.
Our baseline BM25 method scores above the median of submissions in both COLIEE 2020 and 2021 (row 2 vs. 1a).
This confirms that BM25 is a strong baseline and it is in agreement with results from other competitions such as the Health Misinformation and Precision Medicine track of TREC 2020~\cite{pradeep5h2oloo}.

Our pretrained transformer models (rows 3, 4 and 5) score above BM25, the best submission of 2020~\cite{nguyen2020jnlp}, and the second-best team of 2021.
Likewise, our ensemble method effectively combines DeBERTa and monoT5 predictions, achieving the best score among all submissions (row 6). However, the performance of monoT5-zero-shot decreases when combined with DeBERTa (row 5 vs. 7), showing that monoT5-zero-shot is a strong model.

The most interesting comparison is between monoT5 and monoT5-zero-shot (rows 4 and 5).
In the 2020 test data, monoT5 showed better results than monoT5-zero-shot. Hence, we decided to submit only the fine-tuned model to the 2021 competition.
After the release of ground-truth annotations of the 2021 test set, our evaluation of monoT5-zero-shot showed that it performs better than monoT5.

A similar ``inversion'' pattern was found for DeBERTa vs. monoT5 (rows 3 and 4).
DeBERTa was better than monoT5 on the 2020 test set, but the opposite happened on the 2021 test set.

One explanation for these results is that we overfit on the test data of 2020, i.e., by (unintentionally) selecting techniques and hyperparameters that gave the best result on the 2020 test set as experiments progressed. However, this is unlikely to be the case for our fine-tuned monoT5 model, as our hyperparameter selection is fully automatic and maximized on the development set, whose data is from COLIEE competitions before 2020.

Another explanation is that there is a significant difference between the annotation methodologies of 2020 and 2021. Consequently, models specialized in the 2020 data could suffer from this change. However, this is also unlikely since BM25 performed similarly in both years. Furthermore, we cannot confirm this hypothesis since it is difficult to quantify differences in the annotation process.

Regardless of the reason for the inversion, our main finding is that our zero-shot model performed at least comparably to fine-tuned models on the 2020 test set and achieved the best result of a single model on 2021 test data.

\subsection{Ablation of the Answer Selection Method}

In Table~\ref{table:ablation}, we show the ablation result of the answer selection method proposed in Section~\ref{section:answer_selection}.
Our baseline answer selection method, which we refer to as ``no rule'' in the table, uses only the paragraph with the highest score as the final answer set, i.e.,  $\alpha=\gamma=0$ and $\beta=1$.
For all models, the proposed answer selection method gives improvements of 0.6 to two F1 points over the baseline.

\begin{table} 
\centering\centering\resizebox{0.48\textwidth}{!}{
 \begin{tabular}{l | c | c | c | c } 
 \toprule
\textbf{Model} & \textbf{F1} &  \textbf{Prec} &  \textbf{Recall} & $\alpha, \beta, \gamma$  \\
 \midrule 
 monoT5-zero-shot (no rule)  & 0.6517  & 0.7373 & 0.584 & 0, 1, 0\\
 monoT5-zero-shot & 0.6577 & 0.74 & 0.592 & 0, 3, 0.995  \\
 \midrule
 monoT5 (no rule)  & 0.6755 & 0.7600 & 0.608 & 0, 1, 0 \\
 monoT5 & 0.6887 & 0.7155 & 0.6640 & 0, 3, 0.995 \\
 \midrule
 DeBERTa (no rule) & 0.6933 & 0.7800 & 0.6240 & 0, 1, 0  \\ 
 DeBERTa & 0.7094 & 0.7614 & 0.6640  & 0, 2, 0.999 \\ 
 \midrule
 DebertaT5-zero-shot (no rule) & 0.6875 & 0.7777 & 0.6160 & 0, 1, 0 \\
 DebertaT5-zero-shot & 0.7038 & 0.7592 & 0.6560 & 0.6, 2, 0.999 \\
 \midrule
 DebertaT5 (no rule) & 0.7022 & 0.7900 & 0.6320 & 0, 1, 0  \\ 
 DebertaT5  & 0.7217 & 0.7904 & 0.6640 & 0.6, 2, 0.999 \\
 \bottomrule
\end{tabular} 
}
\vspace{0.1cm}
\caption{Ablation on the 2020 data of the answer selection method presented in Section~\ref{section:answer_selection}.}
\label{table:ablation}
\end{table}

\section{Conclusion}

We confirm a counter-intuitive result on a legal case entailment task: that models with little or no adaptation to the target task can have better generalization abilities than models that have been carefully fine-tuned to the task at hand.
Domain adversarial fine-tuning \cite{vernikos2020domain} and changes to the Adam optimizer \cite{chen2020recall} \cite{zhang2021revisiting} have been proposed as valid approaches for fine-tuning Transformer models on small domain-specific datasets. However, whether these techniques could successfully be applied to the legal case entailment task to make models fine-tuned on target task data perform better than zero-shot approaches remains an open question. 

Therefore, although domain-specific language model pretraining and adjustments to the fine-tuning process are promising directions for future research, we believe that zero-shot approaches should not be ignored as strong baselines for such experiments.

It should also be noted that our research has implications for future experiments beyond the scope of legal case entailment tasks. Based on previous work by Yin et al. \cite{yin2020universal, yin2019benchmarking}, it is possible that other legal tasks with limited labeled data, such as legal question answering, may benefit from our zero-shot approach.


\bibliography{acmart}


\begin{thebibliography}{52}


\ifx \showCODEN    \undefined \def \showCODEN     #1{\unskip}     \fi
\ifx \showDOI      \undefined \def \showDOI       #1{#1}\fi
\ifx \showISBNx    \undefined \def \showISBNx     #1{\unskip}     \fi
\ifx \showISBNxiii \undefined \def \showISBNxiii  #1{\unskip}     \fi
\ifx \showISSN     \undefined \def \showISSN      #1{\unskip}     \fi
\ifx \showLCCN     \undefined \def \showLCCN      #1{\unskip}     \fi
\ifx \shownote     \undefined \def \shownote      #1{#1}          \fi
\ifx \showarticletitle \undefined \def \showarticletitle #1{#1}   \fi
\ifx \showURL      \undefined \def \showURL       {\relax}        \fi
\providecommand\bibfield[2]{#2}
\providecommand\bibinfo[2]{#2}
\providecommand\natexlab[1]{#1}
\providecommand\showeprint[2][]{arXiv:#2}

\bibitem[\protect\citeauthoryear{Bajaj, Campos, Craswell, Deng, Gao, Liu,
  Majumder, McNamara, Mitra, Nguyen, Rosenberg, Song, Stoica, Tiwary, and
  Wang}{Bajaj et~al\mbox{.}}{2018}]%
        {MS_MARCO_v3}
\bibfield{author}{\bibinfo{person}{Payal Bajaj}, \bibinfo{person}{Daniel
  Campos}, \bibinfo{person}{Nick Craswell}, \bibinfo{person}{Li Deng},
  \bibinfo{person}{Jianfeng Gao}, \bibinfo{person}{Xiaodong Liu},
  \bibinfo{person}{Rangan Majumder}, \bibinfo{person}{Andrew McNamara},
  \bibinfo{person}{Bhaskar Mitra}, \bibinfo{person}{Tri Nguyen},
  \bibinfo{person}{Mir Rosenberg}, \bibinfo{person}{Xia Song},
  \bibinfo{person}{Alina Stoica}, \bibinfo{person}{Saurabh Tiwary}, {and}
  \bibinfo{person}{Tong Wang}.} \bibinfo{year}{2018}\natexlab{}.
\newblock \showarticletitle{{MS} {MARCO}: {A Human Generated MAchine Reading
  COmprehension Dataset}}.
\newblock \bibinfo{journal}{\emph{arXiv:1611.09268v3}} (\bibinfo{year}{2018}).
\newblock


\bibitem[\protect\citeauthoryear{Bambroo and Awasthi}{Bambroo and
  Awasthi}{2021}]%
        {bambroo2021legaldb}
\bibfield{author}{\bibinfo{person}{Purbid Bambroo} {and} \bibinfo{person}{Aditi
  Awasthi}.} \bibinfo{year}{2021}\natexlab{}.
\newblock \showarticletitle{LegalDB: Long DistilBERT for Legal Document
  Classification}. In \bibinfo{booktitle}{\emph{2021 International Conference
  on Advances in Electrical, Computing, Communication and Sustainable
  Technologies (ICAECT)}}. IEEE, \bibinfo{pages}{1--4}.
\newblock


\bibitem[\protect\citeauthoryear{Bao, Dong, Wei, Wang, Yang, Liu, Wang, Piao,
  Gao, Zhou, and Hon}{Bao et~al\mbox{.}}{2020}]%
        {bao2020unilmv}
\bibfield{author}{\bibinfo{person}{Hangbo Bao}, \bibinfo{person}{Li Dong},
  \bibinfo{person}{Furu Wei}, \bibinfo{person}{Wenhui Wang},
  \bibinfo{person}{Nan Yang}, \bibinfo{person}{Xiaodong Liu},
  \bibinfo{person}{Yu Wang}, \bibinfo{person}{Songhao Piao},
  \bibinfo{person}{Jianfeng Gao}, \bibinfo{person}{Ming Zhou}, {and}
  \bibinfo{person}{Hsiao-Wuen Hon}.} \bibinfo{year}{2020}\natexlab{}.
\newblock \bibinfo{title}{UniLMv2: Pseudo-Masked Language Models for Unified
  Language Model Pre-Training}.
\newblock \bibinfo{howpublished}{ArXiv}.
\newblock
\urldef\tempurl%
\url{https://www.microsoft.com/en-us/research/publication/unilmv2-pseudo-masked-language-models-for-unified-language-model-pre-training/}
\showURL{%
\tempurl}


\bibitem[\protect\citeauthoryear{Brown, Mann, Ryder, Subbiah, Kaplan, Dhariwal,
  Neelakantan, Shyam, Sastry, Askell, Agarwal, Herbert-Voss, Krueger, Henighan,
  Child, Ramesh, Ziegler, Wu, Winter, Hesse, Chen, Sigler, Litwin, Gray, Chess,
  Clark, Berner, McCandlish, Radford, Sutskever, and Amodei}{Brown
  et~al\mbox{.}}{2020}]%
        {NEURIPS2020_1457c0d6}
\bibfield{author}{\bibinfo{person}{Tom Brown}, \bibinfo{person}{Benjamin Mann},
  \bibinfo{person}{Nick Ryder}, \bibinfo{person}{Melanie Subbiah},
  \bibinfo{person}{Jared~D Kaplan}, \bibinfo{person}{Prafulla Dhariwal},
  \bibinfo{person}{Arvind Neelakantan}, \bibinfo{person}{Pranav Shyam},
  \bibinfo{person}{Girish Sastry}, \bibinfo{person}{Amanda Askell},
  \bibinfo{person}{Sandhini Agarwal}, \bibinfo{person}{Ariel Herbert-Voss},
  \bibinfo{person}{Gretchen Krueger}, \bibinfo{person}{Tom Henighan},
  \bibinfo{person}{Rewon Child}, \bibinfo{person}{Aditya Ramesh},
  \bibinfo{person}{Daniel Ziegler}, \bibinfo{person}{Jeffrey Wu},
  \bibinfo{person}{Clemens Winter}, \bibinfo{person}{Chris Hesse},
  \bibinfo{person}{Mark Chen}, \bibinfo{person}{Eric Sigler},
  \bibinfo{person}{Mateusz Litwin}, \bibinfo{person}{Scott Gray},
  \bibinfo{person}{Benjamin Chess}, \bibinfo{person}{Jack Clark},
  \bibinfo{person}{Christopher Berner}, \bibinfo{person}{Sam McCandlish},
  \bibinfo{person}{Alec Radford}, \bibinfo{person}{Ilya Sutskever}, {and}
  \bibinfo{person}{Dario Amodei}.} \bibinfo{year}{2020}\natexlab{}.
\newblock \showarticletitle{Language Models are Few-Shot Learners}. In
  \bibinfo{booktitle}{\emph{Advances in Neural Information Processing
  Systems}}, \bibfield{editor}{\bibinfo{person}{H.~Larochelle},
  \bibinfo{person}{M.~Ranzato}, \bibinfo{person}{R.~Hadsell},
  \bibinfo{person}{M.~F. Balcan}, {and} \bibinfo{person}{H.~Lin}} (Eds.),
  Vol.~\bibinfo{volume}{33}. \bibinfo{publisher}{Curran Associates, Inc.},
  \bibinfo{pages}{1877--1901}.
\newblock
\urldef\tempurl%
\url{https://proceedings.neurips.cc/paper/2020/file/1457c0d6bfcb4967418bfb8ac142f64a-Paper.pdf}
\showURL{%
\tempurl}


\bibitem[\protect\citeauthoryear{Chalkidis and Kampas}{Chalkidis and
  Kampas}{2019}]%
        {Chalkidis2019}
\bibfield{author}{\bibinfo{person}{Ilias Chalkidis} {and}
  \bibinfo{person}{Dimitrios Kampas}.} \bibinfo{year}{2019}\natexlab{}.
\newblock \showarticletitle{Deep learning in law: early adaptation and legal
  word embeddings trained on large corpora}.
\newblock \bibinfo{journal}{\emph{Artificial Intelligence and Law volume 27,
  pages171–198(2019)}} (\bibinfo{year}{2019}).
\newblock


\bibitem[\protect\citeauthoryear{Chen, Hou, Cui, Che, Liu, and Yu}{Chen
  et~al\mbox{.}}{2020}]%
        {chen2020recall}
\bibfield{author}{\bibinfo{person}{Sanyuan Chen}, \bibinfo{person}{Yutai Hou},
  \bibinfo{person}{Yiming Cui}, \bibinfo{person}{Wanxiang Che},
  \bibinfo{person}{Ting Liu}, {and} \bibinfo{person}{Xiangzhan Yu}.}
  \bibinfo{year}{2020}\natexlab{}.
\newblock \bibinfo{title}{Recall and Learn: Fine-tuning Deep Pretrained
  Language Models with Less Forgetting}.
\newblock
\newblock
\showeprint[arxiv]{2004.12651}~[cs.CL]


\bibitem[\protect\citeauthoryear{Devlin, Chang, Lee, and Toutanova}{Devlin
  et~al\mbox{.}}{2019}]%
        {devlin2019bert}
\bibfield{author}{\bibinfo{person}{Jacob Devlin}, \bibinfo{person}{Ming-Wei
  Chang}, \bibinfo{person}{Kenton Lee}, {and} \bibinfo{person}{Kristina
  Toutanova}.} \bibinfo{year}{2019}\natexlab{}.
\newblock \showarticletitle{BERT: Pre-training of Deep Bidirectional
  Transformers for Language Understanding}. In
  \bibinfo{booktitle}{\emph{Proceedings of the 2019 Conference of the North
  American Chapter of the Association for Computational Linguistics: Human
  Language Technologies, Volume 1 (Long and Short Papers)}}.
  \bibinfo{pages}{4171--4186}.
\newblock


\bibitem[\protect\citeauthoryear{Elnaggar, Gebendorfer, Glaser, and
  Matthes}{Elnaggar et~al\mbox{.}}{2018a}]%
        {Elnnagar2}
\bibfield{author}{\bibinfo{person}{Ahmed Elnaggar}, \bibinfo{person}{Christoph
  Gebendorfer}, \bibinfo{person}{Ingo Glaser}, {and} \bibinfo{person}{Florian
  Matthes}.} \bibinfo{year}{2018}\natexlab{a}.
\newblock \showarticletitle{Multi-Task Deep Learning for Legal Document
  Translation, Summarization and Multi-Label Classification}.
\newblock \bibinfo{journal}{\emph{AICCC '18: Proceedings of the 2018 Artificial
  Intelligence and Cloud Computing Conference December 2018 Pages 9–15}}
  (\bibinfo{year}{2018}).
\newblock


\bibitem[\protect\citeauthoryear{Elnaggar, Waltl, Glaser, Landthaler,
  Scepankova, and Matthes}{Elnaggar et~al\mbox{.}}{2018b}]%
        {Elnnagar1}
\bibfield{author}{\bibinfo{person}{Ahmed Elnaggar}, \bibinfo{person}{Bernhard
  Waltl}, \bibinfo{person}{Ingo Glaser}, \bibinfo{person}{Jörg Landthaler},
  \bibinfo{person}{Elena Scepankova}, {and} \bibinfo{person}{Florian Matthes}.}
  \bibinfo{year}{2018}\natexlab{b}.
\newblock \showarticletitle{Stop Illegal Comments: A Multi-Task Deep Learning
  Approach}.
\newblock \bibinfo{journal}{\emph{AICCC '18: Proceedings of the 2018 Artificial
  Intelligence and Cloud Computing Conference December 2018 Pages 41–47}}
  (\bibinfo{year}{2018}).
\newblock


\bibitem[\protect\citeauthoryear{Elwany, Moore, and Oberoi}{Elwany
  et~al\mbox{.}}{2019}]%
        {elwany2019bert}
\bibfield{author}{\bibinfo{person}{Emad Elwany}, \bibinfo{person}{Dave Moore},
  {and} \bibinfo{person}{Gaurav Oberoi}.} \bibinfo{year}{2019}\natexlab{}.
\newblock \showarticletitle{{BERT} Goes to Law School: Quantifying the
  Competitive Advantage of Access to Large Legal Corpora in Contract
  Understanding}. In \bibinfo{booktitle}{\emph{Workshop on Document
  Intelligence at NeurIPS 2019}}.
\newblock


\bibitem[\protect\citeauthoryear{Gao, Dai, and Callan}{Gao
  et~al\mbox{.}}{2021}]%
        {gao2021rethink}
\bibfield{author}{\bibinfo{person}{Luyu Gao}, \bibinfo{person}{Zhuyun Dai},
  {and} \bibinfo{person}{Jamie Callan}.} \bibinfo{year}{2021}\natexlab{}.
\newblock \showarticletitle{Rethink Training of BERT Rerankers in Multi-Stage
  Retrieval Pipeline}.
\newblock \bibinfo{journal}{\emph{arXiv preprint arXiv:2101.08751}}
  (\bibinfo{year}{2021}).
\newblock


\bibitem[\protect\citeauthoryear{He, Liu, Gao, and Chen}{He
  et~al\mbox{.}}{2020}]%
        {he2020deberta}
\bibfield{author}{\bibinfo{person}{Pengcheng He}, \bibinfo{person}{Xiaodong
  Liu}, \bibinfo{person}{Jianfeng Gao}, {and} \bibinfo{person}{Weizhu Chen}.}
  \bibinfo{year}{2020}\natexlab{}.
\newblock \bibinfo{title}{DeBERTa: Decoding-enhanced BERT with Disentangled
  Attention}.
\newblock
\newblock
\showeprint[arxiv]{2006.03654}~[cs.CL]


\bibitem[\protect\citeauthoryear{Kano, Kim, Goebel, and Satoh}{Kano
  et~al\mbox{.}}{2017}]%
        {Kano2017OverviewOC}
\bibfield{author}{\bibinfo{person}{Yoshinobu Kano}, \bibinfo{person}{M. Kim},
  \bibinfo{person}{R. Goebel}, {and} \bibinfo{person}{K. Satoh}.}
  \bibinfo{year}{2017}\natexlab{}.
\newblock \showarticletitle{Overview of {COLIEE} 2017}. In
  \bibinfo{booktitle}{\emph{COLIEE 2017 (EPiC Series in Computing, vol. 47)}}.
  \bibinfo{pages}{1--8}.
\newblock


\bibitem[\protect\citeauthoryear{Kano, Kim, Yoshioka, Lu, Rabelo, Kiyota,
  Goebel, and Satoh}{Kano et~al\mbox{.}}{2018}]%
        {kano2018coliee}
\bibfield{author}{\bibinfo{person}{Yoshinobu Kano}, \bibinfo{person}{Mi-Young
  Kim}, \bibinfo{person}{Masaharu Yoshioka}, \bibinfo{person}{Yao Lu},
  \bibinfo{person}{Juliano Rabelo}, \bibinfo{person}{Naoki Kiyota},
  \bibinfo{person}{Randy Goebel}, {and} \bibinfo{person}{Ken Satoh}.}
  \bibinfo{year}{2018}\natexlab{}.
\newblock \showarticletitle{{COLIEE-2018}: Evaluation of the competition on
  legal information extraction and entailment}. In
  \bibinfo{booktitle}{\emph{JSAI International Symposium on Artificial
  Intelligence}}. \bibinfo{pages}{177--192}.
\newblock


\bibitem[\protect\citeauthoryear{Khashabi, Min, Khot, Sabharwal, Tafjord,
  Clark, and Hajishirzi}{Khashabi et~al\mbox{.}}{2020}]%
        {khashabi2020unifiedqa}
\bibfield{author}{\bibinfo{person}{Daniel Khashabi}, \bibinfo{person}{Sewon
  Min}, \bibinfo{person}{Tushar Khot}, \bibinfo{person}{Ashish Sabharwal},
  \bibinfo{person}{Oyvind Tafjord}, \bibinfo{person}{Peter Clark}, {and}
  \bibinfo{person}{Hannaneh Hajishirzi}.} \bibinfo{year}{2020}\natexlab{}.
\newblock \showarticletitle{UnifiedQA: Crossing Format Boundaries With a Single
  QA System}. In \bibinfo{booktitle}{\emph{Proceedings of the 2020 Conference
  on Empirical Methods in Natural Language Processing: Findings}}.
  \bibinfo{pages}{1896--1907}.
\newblock


\bibitem[\protect\citeauthoryear{Leivaditi, Rossi, and Kanoulas}{Leivaditi
  et~al\mbox{.}}{2020}]%
        {leivaditi2020benchmark}
\bibfield{author}{\bibinfo{person}{Spyretta Leivaditi}, \bibinfo{person}{Julien
  Rossi}, {and} \bibinfo{person}{Evangelos Kanoulas}.}
  \bibinfo{year}{2020}\natexlab{}.
\newblock \showarticletitle{A Benchmark for Lease Contract Review}.
\newblock \bibinfo{journal}{\emph{arXiv preprint arXiv:2010.10386}}
  (\bibinfo{year}{2020}).
\newblock


\bibitem[\protect\citeauthoryear{Lewis, Liu, Goyal, Ghazvininejad, Mohamed,
  Levy, Stoyanov, and Zettlemoyer}{Lewis et~al\mbox{.}}{2020}]%
        {lewis2020bart}
\bibfield{author}{\bibinfo{person}{Mike Lewis}, \bibinfo{person}{Yinhan Liu},
  \bibinfo{person}{Naman Goyal}, \bibinfo{person}{Marjan Ghazvininejad},
  \bibinfo{person}{Abdelrahman Mohamed}, \bibinfo{person}{Omer Levy},
  \bibinfo{person}{Veselin Stoyanov}, {and} \bibinfo{person}{Luke
  Zettlemoyer}.} \bibinfo{year}{2020}\natexlab{}.
\newblock \showarticletitle{BART: Denoising Sequence-to-Sequence Pre-training
  for Natural Language Generation, Translation, and Comprehension}. In
  \bibinfo{booktitle}{\emph{Proceedings of the 58th Annual Meeting of the
  Association for Computational Linguistics}}. \bibinfo{pages}{7871--7880}.
\newblock


\bibitem[\protect\citeauthoryear{Lin, Ma, Lin, Yang, Pradeep, and Nogueira}{Lin
  et~al\mbox{.}}{2021}]%
        {lin2021pyserini}
\bibfield{author}{\bibinfo{person}{Jimmy Lin}, \bibinfo{person}{Xueguang Ma},
  \bibinfo{person}{Sheng-Chieh Lin}, \bibinfo{person}{Jheng-Hong Yang},
  \bibinfo{person}{Ronak Pradeep}, {and} \bibinfo{person}{Rodrigo Nogueira}.}
  \bibinfo{year}{2021}\natexlab{}.
\newblock \showarticletitle{Pyserini: An Easy-to-Use Python Toolkit to Support
  Replicable IR Research with Sparse and Dense Representations}.
\newblock \bibinfo{journal}{\emph{arXiv preprint arXiv:2102.10073}}
  (\bibinfo{year}{2021}).
\newblock


\bibitem[\protect\citeauthoryear{Lin, Nogueira, and Yates}{Lin
  et~al\mbox{.}}{2020}]%
        {lin2020pretrained}
\bibfield{author}{\bibinfo{person}{Jimmy Lin}, \bibinfo{person}{Rodrigo
  Nogueira}, {and} \bibinfo{person}{Andrew Yates}.}
  \bibinfo{year}{2020}\natexlab{}.
\newblock \showarticletitle{Pretrained transformers for text ranking: Bert and
  beyond}.
\newblock \bibinfo{journal}{\emph{arXiv preprint arXiv:2010.06467}}
  (\bibinfo{year}{2020}).
\newblock


\bibitem[\protect\citeauthoryear{Lu, Gong, Ye, and Zhang}{Lu
  et~al\mbox{.}}{2020}]%
        {lu2020learning}
\bibfield{author}{\bibinfo{person}{Jiang Lu}, \bibinfo{person}{Pinghua Gong},
  \bibinfo{person}{Jieping Ye}, {and} \bibinfo{person}{Changshui Zhang}.}
  \bibinfo{year}{2020}\natexlab{}.
\newblock \showarticletitle{Learning from Very Few Samples: A Survey}.
\newblock \bibinfo{journal}{\emph{arXiv preprint arXiv:2009.02653}}
  (\bibinfo{year}{2020}).
\newblock


\bibitem[\protect\citeauthoryear{Ma, Guo, Zhang, Fan, Ji, and Cheng}{Ma
  et~al\mbox{.}}{2021}]%
        {ma2021prop}
\bibfield{author}{\bibinfo{person}{Xinyu Ma}, \bibinfo{person}{Jiafeng Guo},
  \bibinfo{person}{Ruqing Zhang}, \bibinfo{person}{Yixing Fan},
  \bibinfo{person}{Xiang Ji}, {and} \bibinfo{person}{Xueqi Cheng}.}
  \bibinfo{year}{2021}\natexlab{}.
\newblock \showarticletitle{PROP: Pre-training with Representative Words
  Prediction for Ad-hoc Retrieval}. In \bibinfo{booktitle}{\emph{Proceedings of
  the 14th ACM International Conference on Web Search and Data Mining}}.
  \bibinfo{pages}{283--291}.
\newblock


\bibitem[\protect\citeauthoryear{Mikolov}{Mikolov}{2013}]%
        {word2vec2}
\bibfield{author}{\bibinfo{person}{Tomas Mikolov}.}
  \bibinfo{year}{2013}\natexlab{}.
\newblock \showarticletitle{Distributed representations of words and phrases
  and their compositionality}.
\newblock \bibinfo{journal}{\emph{arXiv preprint arXiv:1310.4546}}
  (\bibinfo{year}{2013}).
\newblock


\bibitem[\protect\citeauthoryear{Mikolov, Chen, Corrado, and Dean}{Mikolov
  et~al\mbox{.}}{2013}]%
        {word2vec1}
\bibfield{author}{\bibinfo{person}{Tomas Mikolov}, \bibinfo{person}{Kai Chen},
  \bibinfo{person}{Greg Corrado}, {and} \bibinfo{person}{Jeffrey Dean}.}
  \bibinfo{year}{2013}\natexlab{}.
\newblock \showarticletitle{Efficient Estimation of Word Representations in
  Vector Space}.
\newblock \bibinfo{journal}{\emph{arXiv preprint arXiv:1301.3781}}
  (\bibinfo{year}{2013}).
\newblock


\bibitem[\protect\citeauthoryear{Nguyen, Vuong, Nguyen, Dang, Bui, Vu, Nguyen,
  Tran, Satoh, and Nguyen}{Nguyen et~al\mbox{.}}{2020}]%
        {nguyen2020jnlp}
\bibfield{author}{\bibinfo{person}{Ha-Thanh Nguyen},
  \bibinfo{person}{Hai-Yen~Thi Vuong}, \bibinfo{person}{Phuong~Minh Nguyen},
  \bibinfo{person}{Binh~Tran Dang}, \bibinfo{person}{Quan~Minh Bui},
  \bibinfo{person}{Sinh~Trong Vu}, \bibinfo{person}{Chau~Minh Nguyen},
  \bibinfo{person}{Vu Tran}, \bibinfo{person}{Ken Satoh}, {and}
  \bibinfo{person}{Minh~Le Nguyen}.} \bibinfo{year}{2020}\natexlab{}.
\newblock \showarticletitle{JNLP Team: Deep Learning for Legal Processing in
  COLIEE 2020}.
\newblock \bibinfo{journal}{\emph{arXiv preprint arXiv:2011.08071}}
  (\bibinfo{year}{2020}).
\newblock


\bibitem[\protect\citeauthoryear{Nogueira, Jiang, Pradeep, and Lin}{Nogueira
  et~al\mbox{.}}{2020}]%
        {nogueira2020document}
\bibfield{author}{\bibinfo{person}{Rodrigo Nogueira}, \bibinfo{person}{Zhiying
  Jiang}, \bibinfo{person}{Ronak Pradeep}, {and} \bibinfo{person}{Jimmy Lin}.}
  \bibinfo{year}{2020}\natexlab{}.
\newblock \showarticletitle{Document Ranking with a Pretrained
  Sequence-to-Sequence Model}. In \bibinfo{booktitle}{\emph{Proceedings of the
  2020 Conference on Empirical Methods in Natural Language Processing:
  Findings}}. \bibinfo{pages}{708--718}.
\newblock


\bibitem[\protect\citeauthoryear{Peters, Ruder, and Smith}{Peters
  et~al\mbox{.}}{2019}]%
        {peters2019tune}
\bibfield{author}{\bibinfo{person}{Matthew~E Peters},
  \bibinfo{person}{Sebastian Ruder}, {and} \bibinfo{person}{Noah~A Smith}.}
  \bibinfo{year}{2019}\natexlab{}.
\newblock \showarticletitle{To Tune or Not to Tune? Adapting Pretrained
  Representations to Diverse Tasks}. In \bibinfo{booktitle}{\emph{Proceedings
  of the 4th Workshop on Representation Learning for NLP (RepL4NLP-2019)}}.
  \bibinfo{pages}{7--14}.
\newblock


\bibitem[\protect\citeauthoryear{Pradeep, Ma, Zhang, Cui, Xu, Nogueira, and
  Lin}{Pradeep et~al\mbox{.}}{[n.d.]}]%
        {pradeep5h2oloo}
\bibfield{author}{\bibinfo{person}{Ronak Pradeep}, \bibinfo{person}{Xueguang
  Ma}, \bibinfo{person}{Xinyu Zhang}, \bibinfo{person}{Hang Cui},
  \bibinfo{person}{Ruizhou Xu}, \bibinfo{person}{Rodrigo Nogueira}, {and}
  \bibinfo{person}{Jimmy Lin}.} \bibinfo{year}{[n.d.]}\natexlab{}.
\newblock \showarticletitle{H2oloo at TREC 2020: When all you got is a
  hammer... Deep Learning, Health Misinformation, and Precision Medicine}.
\newblock \bibinfo{journal}{\emph{Corpus}} \bibinfo{volume}{5},
  \bibinfo{number}{d3} (\bibinfo{year}{[n.\,d.]}), \bibinfo{pages}{d2}.
\newblock


\bibitem[\protect\citeauthoryear{Rabelo, Kim, and Goebel}{Rabelo
  et~al\mbox{.}}{2020a}]%
        {rabelo2020}
\bibfield{author}{\bibinfo{person}{J. Rabelo}, \bibinfo{person}{M.Y. Kim},
  {and} \bibinfo{person}{R. Goebel}.} \bibinfo{year}{2020}\natexlab{a}.
\newblock \showarticletitle{Application of text entailment techniques in COLIEE
  2020}.
\newblock \bibinfo{journal}{\emph{International Workshop on Juris-informatics
  (JURISIN) associated with JSAI International Symposia on AI (JSAI-isAI)}}
  (\bibinfo{year}{2020}).
\newblock


\bibitem[\protect\citeauthoryear{Rabelo, Kim, and Goebel}{Rabelo
  et~al\mbox{.}}{2019a}]%
        {rabelo2019combining}
\bibfield{author}{\bibinfo{person}{Juliano Rabelo}, \bibinfo{person}{Mi-Young
  Kim}, {and} \bibinfo{person}{Randy Goebel}.}
  \bibinfo{year}{2019}\natexlab{a}.
\newblock \showarticletitle{Combining similarity and transformer methods for
  case law entailment}. In \bibinfo{booktitle}{\emph{Proceedings of the
  Seventeenth International Conference on Artificial Intelligence and Law
  (ICAIL '19)}}. \bibinfo{pages}{290--296}.
\newblock


\bibitem[\protect\citeauthoryear{Rabelo, Kim, Goebel, Yoshioka, Kano, and
  Satoh}{Rabelo et~al\mbox{.}}{2019b}]%
        {rabelo2019coliee}
\bibfield{author}{\bibinfo{person}{Juliano Rabelo}, \bibinfo{person}{Mi-Young
  Kim}, \bibinfo{person}{Randy Goebel}, \bibinfo{person}{Masaharu Yoshioka},
  \bibinfo{person}{Yoshinobu Kano}, {and} \bibinfo{person}{Ken Satoh}.}
  \bibinfo{year}{2019}\natexlab{b}.
\newblock \showarticletitle{A Summary of the {COLIEE} 2019 Competition}. In
  \bibinfo{booktitle}{\emph{JSAI International Symposium on Artificial
  Intelligence}}. \bibinfo{pages}{34--49}.
\newblock


\bibitem[\protect\citeauthoryear{Rabelo, Kim, Goebel, Yoshioka, Kano, and
  Satoh}{Rabelo et~al\mbox{.}}{2020b}]%
        {rabelo2020coliee}
\bibfield{author}{\bibinfo{person}{Juliano Rabelo}, \bibinfo{person}{Mi-Young
  Kim}, \bibinfo{person}{Randy Goebel}, \bibinfo{person}{Masaharu Yoshioka},
  \bibinfo{person}{Yoshinobu Kano}, {and} \bibinfo{person}{Ken Satoh}.}
  \bibinfo{year}{2020}\natexlab{b}.
\newblock \showarticletitle{COLIEE 2020: Methods for Legal Document Retrieval
  and Entailment}.
\newblock  (\bibinfo{year}{2020}).
\newblock


\bibitem[\protect\citeauthoryear{Radford, Kim, Hallacy, Ramesh, Goh, Agarwal,
  Sastry, Askell, Mishkin, Clark, et~al\mbox{.}}{Radford et~al\mbox{.}}{2021}]%
        {radford2021learning}
\bibfield{author}{\bibinfo{person}{Alec Radford}, \bibinfo{person}{Jong~Wook
  Kim}, \bibinfo{person}{Chris Hallacy}, \bibinfo{person}{Aditya Ramesh},
  \bibinfo{person}{Gabriel Goh}, \bibinfo{person}{Sandhini Agarwal},
  \bibinfo{person}{Girish Sastry}, \bibinfo{person}{Amanda Askell},
  \bibinfo{person}{Pamela Mishkin}, \bibinfo{person}{Jack Clark},
  {et~al\mbox{.}}} \bibinfo{year}{2021}\natexlab{}.
\newblock \showarticletitle{Learning transferable visual models from natural
  language supervision}.
\newblock \bibinfo{journal}{\emph{arXiv preprint arXiv:2103.00020}}
  (\bibinfo{year}{2021}).
\newblock


\bibitem[\protect\citeauthoryear{Raffel, Shazeer, Roberts, Lee, Narang, Matena,
  Zhou, Li, and Liu}{Raffel et~al\mbox{.}}{2020}]%
        {raffel2020t5}
\bibfield{author}{\bibinfo{person}{Colin Raffel}, \bibinfo{person}{Noam
  Shazeer}, \bibinfo{person}{Adam Roberts}, \bibinfo{person}{Katherine Lee},
  \bibinfo{person}{Sharan Narang}, \bibinfo{person}{Michael Matena},
  \bibinfo{person}{Yanqi Zhou}, \bibinfo{person}{Wei Li}, {and}
  \bibinfo{person}{Peter~J. Liu}.} \bibinfo{year}{2020}\natexlab{}.
\newblock \showarticletitle{Exploring the Limits of Transfer Learning with a
  Unified Text-to-Text Transformer}.
\newblock \bibinfo{journal}{\emph{Journal of Machine Learning Research}}
  \bibinfo{volume}{21}, \bibinfo{number}{140} (\bibinfo{year}{2020}),
  \bibinfo{pages}{1--67}.
\newblock
\urldef\tempurl%
\url{http://jmlr.org/papers/v21/20-074.html}
\showURL{%
\tempurl}


\bibitem[\protect\citeauthoryear{Roberts, Demner-Fushman, Voorhees, Hersh,
  Bedrick, Lazar, and Pant}{Roberts et~al\mbox{.}}{2019}]%
        {Roberts2019OverviewOT}
\bibfield{author}{\bibinfo{person}{Kirk Roberts}, \bibinfo{person}{Dina
  Demner-Fushman}, \bibinfo{person}{E. Voorhees}, \bibinfo{person}{W. Hersh},
  \bibinfo{person}{Steven Bedrick}, \bibinfo{person}{Alexander~J. Lazar}, {and}
  \bibinfo{person}{S. Pant}.} \bibinfo{year}{2019}\natexlab{}.
\newblock \showarticletitle{Overview of the TREC 2019 Precision Medicine
  Track}.
\newblock \bibinfo{journal}{\emph{The ... text REtrieval conference : TREC.
  Text REtrieval Conference}}  \bibinfo{volume}{26} (\bibinfo{year}{2019}).
\newblock


\bibitem[\protect\citeauthoryear{Robertson, Walker, Jones, Hancock-Beaulieu,
  Gatford, et~al\mbox{.}}{Robertson et~al\mbox{.}}{1995}]%
        {robertson1995okapi}
\bibfield{author}{\bibinfo{person}{Stephen~E Robertson}, \bibinfo{person}{Steve
  Walker}, \bibinfo{person}{Susan Jones}, \bibinfo{person}{Micheline~M
  Hancock-Beaulieu}, \bibinfo{person}{Mike Gatford}, {et~al\mbox{.}}}
  \bibinfo{year}{1995}\natexlab{}.
\newblock \showarticletitle{Okapi at TREC-3}.
\newblock \bibinfo{journal}{\emph{Nist Special Publication Sp}}
  \bibinfo{volume}{109} (\bibinfo{year}{1995}), \bibinfo{pages}{109}.
\newblock


\bibitem[\protect\citeauthoryear{Schick and Sch{\"u}tze}{Schick and
  Sch{\"u}tze}{2020}]%
        {schick2020exploiting}
\bibfield{author}{\bibinfo{person}{Timo Schick} {and} \bibinfo{person}{Hinrich
  Sch{\"u}tze}.} \bibinfo{year}{2020}\natexlab{}.
\newblock \showarticletitle{Exploiting cloze questions for few-shot text
  classification and natural language inference}.
\newblock \bibinfo{journal}{\emph{arXiv preprint arXiv:2001.07676}}
  (\bibinfo{year}{2020}).
\newblock


\bibitem[\protect\citeauthoryear{Shaghaghian, Feng, Jafarpour, and
  Pogrebnyakov}{Shaghaghian et~al\mbox{.}}{2020}]%
        {shaghaghian2020customizing}
\bibfield{author}{\bibinfo{person}{Shohreh Shaghaghian},
  \bibinfo{person}{Luna~Yue Feng}, \bibinfo{person}{Borna Jafarpour}, {and}
  \bibinfo{person}{Nicolai Pogrebnyakov}.} \bibinfo{year}{2020}\natexlab{}.
\newblock \showarticletitle{Customizing Contextualized Language Models for
  Legal Document Reviews}. In \bibinfo{booktitle}{\emph{2020 IEEE International
  Conference on Big Data (Big Data)}}. IEEE, \bibinfo{pages}{2139--2148}.
\newblock


\bibitem[\protect\citeauthoryear{Tam, Menon, Bansal, Srivastava, and
  Raffel}{Tam et~al\mbox{.}}{2021}]%
        {tam2021improving}
\bibfield{author}{\bibinfo{person}{Derek Tam}, \bibinfo{person}{Rakesh~R
  Menon}, \bibinfo{person}{Mohit Bansal}, \bibinfo{person}{Shashank
  Srivastava}, {and} \bibinfo{person}{Colin Raffel}.}
  \bibinfo{year}{2021}\natexlab{}.
\newblock \showarticletitle{Improving and Simplifying Pattern Exploiting
  Training}.
\newblock \bibinfo{journal}{\emph{arXiv preprint arXiv:2103.11955}}
  (\bibinfo{year}{2021}).
\newblock


\bibitem[\protect\citeauthoryear{Thakur, Reimers, Rücklé, Srivastava, and
  Gurevych}{Thakur et~al\mbox{.}}{2021}]%
        {thakur2021beir}
\bibfield{author}{\bibinfo{person}{Nandan Thakur}, \bibinfo{person}{Nils
  Reimers}, \bibinfo{person}{Andreas Rücklé}, \bibinfo{person}{Abhishek
  Srivastava}, {and} \bibinfo{person}{Iryna Gurevych}.}
  \bibinfo{year}{2021}\natexlab{}.
\newblock \showarticletitle{BEIR: A Heterogenous Benchmark for Zero-shot
  Evaluation of Information Retrieval Models}.
\newblock \bibinfo{journal}{\emph{arXiv preprint arXiv:2104.08663}}
  (\bibinfo{date}{4} \bibinfo{year}{2021}).
\newblock
\urldef\tempurl%
\url{https://arxiv.org/abs/2104.08663}
\showURL{%
\tempurl}


\bibitem[\protect\citeauthoryear{Vaswani, Shazeer, Parmar, Uszkoreit, Jones,
  Gomez, Kaiser, and Polosukhin}{Vaswani et~al\mbox{.}}{2017}]%
        {vaswani2017attention}
\bibfield{author}{\bibinfo{person}{Ashish Vaswani}, \bibinfo{person}{Noam
  Shazeer}, \bibinfo{person}{Niki Parmar}, \bibinfo{person}{Jakob Uszkoreit},
  \bibinfo{person}{Llion Jones}, \bibinfo{person}{Aidan~N Gomez},
  \bibinfo{person}{Lukasz Kaiser}, {and} \bibinfo{person}{Illia Polosukhin}.}
  \bibinfo{year}{2017}\natexlab{}.
\newblock \showarticletitle{Attention is All you Need}. In
  \bibinfo{booktitle}{\emph{NIPS}}.
\newblock


\bibitem[\protect\citeauthoryear{Vernikos, Margatina, Chronopoulou, and
  Androutsopoulos}{Vernikos et~al\mbox{.}}{2020}]%
        {vernikos2020domain}
\bibfield{author}{\bibinfo{person}{Giorgos Vernikos}, \bibinfo{person}{Katerina
  Margatina}, \bibinfo{person}{Alexandra Chronopoulou}, {and}
  \bibinfo{person}{Ion Androutsopoulos}.} \bibinfo{year}{2020}\natexlab{}.
\newblock \bibinfo{title}{Domain Adversarial Fine-Tuning as an Effective
  Regularizer}.
\newblock
\newblock
\showeprint[arxiv]{2009.13366}~[cs.LG]


\bibitem[\protect\citeauthoryear{Voorhees}{Voorhees}{2004}]%
        {trec2004}
\bibfield{author}{\bibinfo{person}{Ellen~M. Voorhees}.}
  \bibinfo{year}{2004}\natexlab{}.
\newblock \showarticletitle{Overview of the TREC 2004 Robust Track}.
\newblock \bibinfo{journal}{\emph{Proceedings of the Thirteenth Text REtrieval
  Conference, TREC 2004, Gaithersburg, Maryland, November 16-19, 2004}}
  (\bibinfo{year}{2004}).
\newblock


\bibitem[\protect\citeauthoryear{Xiao, Zhong, Guo, Tu, Liu, Sun, Feng, Han, Hu,
  Wang, and Xu}{Xiao et~al\mbox{.}}{2018}]%
        {xiao2018cail2018}
\bibfield{author}{\bibinfo{person}{Chaojun Xiao}, \bibinfo{person}{Haoxi
  Zhong}, \bibinfo{person}{Zhipeng Guo}, \bibinfo{person}{Cunchao Tu},
  \bibinfo{person}{Zhiyuan Liu}, \bibinfo{person}{Maosong Sun},
  \bibinfo{person}{Yansong Feng}, \bibinfo{person}{Xianpei Han},
  \bibinfo{person}{Zhen Hu}, \bibinfo{person}{Heng Wang}, {and}
  \bibinfo{person}{Jianfeng Xu}.} \bibinfo{year}{2018}\natexlab{}.
\newblock \showarticletitle{{CAIL2018}: A Large-Scale Legal Dataset for
  Judgment Prediction}.
\newblock \bibinfo{journal}{\emph{arXiv:1807.02478}} (\bibinfo{year}{2018}).
\newblock


\bibitem[\protect\citeauthoryear{Yeung}{Yeung}{2019}]%
        {yeung2019effects}
\bibfield{author}{\bibinfo{person}{Chin~Man Yeung}.}
  \bibinfo{year}{2019}\natexlab{}.
\newblock \emph{\bibinfo{title}{Effects of inserting domain vocabulary and
  fine-tuning {BERT} for {German} legal language}}.
\newblock \bibinfo{thesistype}{Master's\ thesis}. \bibinfo{school}{University
  of Twente}.
\newblock


\bibitem[\protect\citeauthoryear{Yin, Hay, and Roth}{Yin et~al\mbox{.}}{2019}]%
        {yin2019benchmarking}
\bibfield{author}{\bibinfo{person}{Wenpeng Yin}, \bibinfo{person}{Jamaal Hay},
  {and} \bibinfo{person}{Dan Roth}.} \bibinfo{year}{2019}\natexlab{}.
\newblock \bibinfo{title}{Benchmarking Zero-shot Text Classification: Datasets,
  Evaluation and Entailment Approach}.
\newblock
\newblock
\showeprint[arxiv]{1909.00161}~[cs.CL]


\bibitem[\protect\citeauthoryear{Yin, Rajani, Radev, Socher, and Xiong}{Yin
  et~al\mbox{.}}{2020}]%
        {yin2020universal}
\bibfield{author}{\bibinfo{person}{Wenpeng Yin},
  \bibinfo{person}{Nazneen~Fatema Rajani}, \bibinfo{person}{Dragomir Radev},
  \bibinfo{person}{Richard Socher}, {and} \bibinfo{person}{Caiming Xiong}.}
  \bibinfo{year}{2020}\natexlab{}.
\newblock \bibinfo{title}{Universal Natural Language Processing with Limited
  Annotations: Try Few-shot Textual Entailment as a Start}.
\newblock
\newblock
\showeprint[arxiv]{2010.02584}~[cs.CL]


\bibitem[\protect\citeauthoryear{Yin, Sch{\"u}tze, Xiang, and Zhou}{Yin
  et~al\mbox{.}}{2016}]%
        {yin2016abcnn}
\bibfield{author}{\bibinfo{person}{Wenpeng Yin}, \bibinfo{person}{Hinrich
  Sch{\"u}tze}, \bibinfo{person}{Bing Xiang}, {and} \bibinfo{person}{Bowen
  Zhou}.} \bibinfo{year}{2016}\natexlab{}.
\newblock \showarticletitle{{ABCNN}: Attention-based convolutional neural
  network for modeling sentence pairs}.
\newblock \bibinfo{journal}{\emph{Transactions of the Association for
  Computational Linguistics}}  \bibinfo{volume}{4} (\bibinfo{year}{2016}),
  \bibinfo{pages}{259--272}.
\newblock


\bibitem[\protect\citeauthoryear{Zhang, Gupta, Nogueira, Cho, and Lin}{Zhang
  et~al\mbox{.}}{2020}]%
        {zhang2020rapidly}
\bibfield{author}{\bibinfo{person}{Edwin Zhang}, \bibinfo{person}{Nikhil
  Gupta}, \bibinfo{person}{Rodrigo Nogueira}, \bibinfo{person}{Kyunghyun Cho},
  {and} \bibinfo{person}{Jimmy Lin}.} \bibinfo{year}{2020}\natexlab{}.
\newblock \showarticletitle{Rapidly Deploying a Neural Search Engine for the
  COVID-19 Open Research Dataset}. In \bibinfo{booktitle}{\emph{Proceedings of
  the 1st Workshop on NLP for COVID-19 at ACL 2020}}.
\newblock


\bibitem[\protect\citeauthoryear{Zhang, Wu, Katiyar, Weinberger, and
  Artzi}{Zhang et~al\mbox{.}}{2021}]%
        {zhang2021revisiting}
\bibfield{author}{\bibinfo{person}{Tianyi Zhang}, \bibinfo{person}{Felix Wu},
  \bibinfo{person}{Arzoo Katiyar}, \bibinfo{person}{Kilian~Q. Weinberger},
  {and} \bibinfo{person}{Yoav Artzi}.} \bibinfo{year}{2021}\natexlab{}.
\newblock \bibinfo{title}{Revisiting Few-sample BERT Fine-tuning}.
\newblock
\newblock
\showeprint[arxiv]{2006.05987}~[cs.CL]


\bibitem[\protect\citeauthoryear{Zhong, Xiao, Tu, Zhang, Liu, and Sun}{Zhong
  et~al\mbox{.}}{2020b}]%
        {zhong2020}
\bibfield{author}{\bibinfo{person}{Haoxi Zhong}, \bibinfo{person}{Chaojun
  Xiao}, \bibinfo{person}{Cunchao Tu}, \bibinfo{person}{Tianyang Zhang},
  \bibinfo{person}{Zhiyuan Liu}, {and} \bibinfo{person}{Maosong Sun}.}
  \bibinfo{year}{2020}\natexlab{b}.
\newblock \showarticletitle{How Does NLP Benefit Legal System: A Summary of
  Legal Artificial Intelligence}.
\newblock \bibinfo{journal}{\emph{arXiv preprint arXiv:2004.12158}}
  (\bibinfo{year}{2020}).
\newblock


\bibitem[\protect\citeauthoryear{Zhong, Xiao, Tu, Zhang, Liu, and Sun}{Zhong
  et~al\mbox{.}}{2020c}]%
        {zhong2020does}
\bibfield{author}{\bibinfo{person}{Haoxi Zhong}, \bibinfo{person}{Chaojun
  Xiao}, \bibinfo{person}{Cunchao Tu}, \bibinfo{person}{Tianyang Zhang},
  \bibinfo{person}{Zhiyuan Liu}, {and} \bibinfo{person}{Maosong Sun}.}
  \bibinfo{year}{2020}\natexlab{c}.
\newblock \showarticletitle{How Does {NLP} Benefit Legal System: A Summary of
  Legal Artificial Intelligence}.
\newblock \bibinfo{journal}{\emph{arXiv:2004.12158}} (\bibinfo{year}{2020}).
\newblock


\bibitem[\protect\citeauthoryear{Zhong, Xiao, Tu, Zhang, Liu, and Sun1}{Zhong
  et~al\mbox{.}}{2020a}]%
        {zhong2020b}
\bibfield{author}{\bibinfo{person}{Haoxi Zhong}, \bibinfo{person}{Chaojun
  Xiao}, \bibinfo{person}{Cunchao Tu}, \bibinfo{person}{Tianyang Zhang},
  \bibinfo{person}{Zhiyuan Liu}, {and} \bibinfo{person}{Maosong Sun1}.}
  \bibinfo{year}{2020}\natexlab{a}.
\newblock \showarticletitle{JEC-QA: A Legal-Domain Question Answering Dataset.}
\newblock \bibinfo{journal}{\emph{Proceedings of the AAAI Conference on
  Artificial Intelligence, 34(05), 9701-9708.}} (\bibinfo{year}{2020}).
\newblock


\end{thebibliography}
\bibliographystyle{ACM-Reference-Format}


\end{document}